\documentclass[10pt,twocolumn,letterpaper]{article}

\usepackage{cvpr}              

\usepackage{graphicx}
\usepackage{amsmath}
\usepackage{amssymb}
\usepackage{booktabs}

\usepackage{array}
\usepackage{textcomp}
\usepackage{stfloats}
\usepackage{url}
\usepackage{verbatim}
\usepackage{graphicx}
\usepackage{cite}
\usepackage{multirow}
\usepackage{setspace}
\usepackage{xcolor}
\usepackage[accsupp]{axessibility}
\usepackage[pagebackref,breaklinks,colorlinks]{hyperref}

\usepackage[capitalize]{cleveref}
\crefname{section}{Sec.}{Secs.}
\Crefname{section}{Section}{Sections}
\Crefname{table}{Table}{Tables}
\crefname{table}{Tab.}{Tabs.}

\def\cvprPaperID{3132} 
\def\confName{CVPR}
\def\confYear{2023}

\begin{document}

\title{TranSG: Transformer-Based Skeleton Graph Prototype Contrastive Learning with Structure-Trajectory Prompted Reconstruction for Person Re-Identification}

\author{Haocong Rao\quad Chunyan Miao\thanks{Corresponding author}\\
LILY Research Center, Nanyang Technological University, Singapore\\
School of Computer Science and Engineering, Nanyang Technological University, Singapore\\
{\tt\small \{haocong001, ascymiao\}@ntu.edu.sg}
\renewcommand{\thefootnote}{\fnsymbol{footnote}}
}
\maketitle
\begin{abstract}
Person re-identification (re-ID) via 3D skeleton data is an emerging topic with prominent advantages. Existing methods usually design skeleton descriptors with raw body joints or perform skeleton sequence representation learning. However, they typically cannot concurrently model different body-component relations, and rarely explore useful semantics from fine-grained representations of body joints. In this paper, we propose a generic Transformer-based Skeleton Graph prototype contrastive learning (TranSG) approach with structure-trajectory prompted reconstruction to fully capture skeletal relations and valuable spatial-temporal semantics from skeleton graphs for person re-ID. Specifically, we first devise the Skeleton Graph Transformer (SGT) to simultaneously learn body and motion relations within skeleton graphs, so as to aggregate key correlative node features into graph representations. Then, we propose the Graph Prototype Contrastive learning (GPC) to mine the most typical graph features (graph prototypes) of each identity, and contrast the inherent similarity between graph representations and different prototypes from both skeleton and sequence levels to learn discriminative graph representations. Last, a graph Structure-Trajectory Prompted Reconstruction (STPR) mechanism is proposed to exploit the spatial and temporal contexts of graph nodes to prompt skeleton graph reconstruction, which facilitates capturing more valuable patterns and graph semantics for person re-ID. 
Empirical evaluations demonstrate that TranSG significantly outperforms existing state-of-the-art methods. We further show its generality under different graph modeling, RGB-estimated skeletons, and unsupervised scenarios. 
Our codes are available at \href{https://github.com/Kali-Hac/TranSG}{https://github.com/Kali-Hac/TranSG}.



\end{abstract}

\section{Introduction}

Person re-identification (re-ID) is a challenging task of retrieving and matching a specific person across varying views or scenarios, which has empowered many vital applications such as security authentication, human tracking, and robotics \cite{ye2021deep,rao2021self,nambiar2019gait,vezzani2013people}. Recently driven by economical, non-obtrusive and accurate skeleton-tracking devices like Kinect \cite{shotton2011real-time}, person re-ID via 3D skeletons has attracted surging attention in both academia and industry \cite{rao2021self,rao2020self,pala2019enhanced,liao2020model,andersson2015person,munaro20143d,rao2022skeleton,munaro2014one,barbosa2012re,rao2022simmc}. Unlike conventional methods that rely on visual appearance features ($e.g.,$ colors, silhouettes), skeleton-based person re-ID methods model unique body and motion representations with 3D positions of key body joints, which typically enjoy smaller data sizes, lighter models, and more robust performance under scale and view variations \cite{han2017space,rao2021self}.

\begin{figure}
    \centering
    \scalebox{0.55}{
    \includegraphics{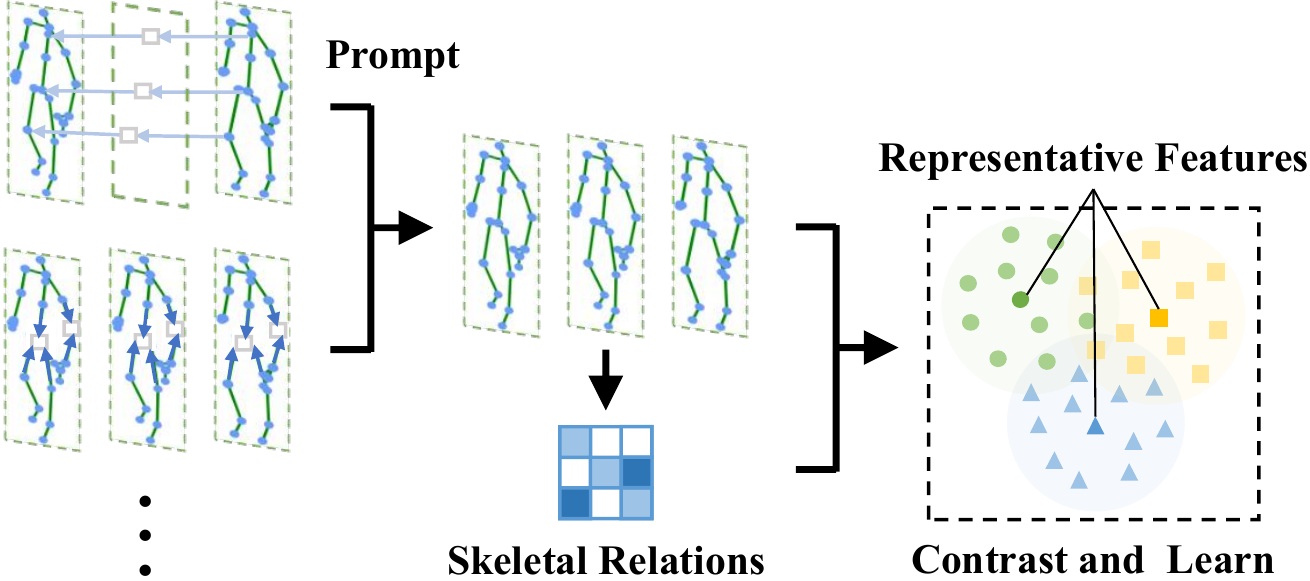}
    }
    \caption{TranSG temporally and spatially masks skeleton graphs to prompt their reconstruction, while integrating full skeletal relations into contrastive learning of typical features for person re-ID.}
    \label{first}
\end{figure}

Most existing methods manually extract anthropometric descriptors and gait attributes from body-joint coordinates \cite{pala2019enhanced,liao2020model,andersson2015person,munaro20143d,barbosa2012re}, or leverage sequence learning paradigms such as Long Short-Term Memory (LSTM) \cite{hochreiter1997long} to model body and motion features with skeleton sequences \cite{haque2016recurrent,rao2021self,rao2020self,rao2022simmc}. However, these methods rarely explore the inherent body relations within skeletons ($e.g.,$ inter-joint motion correlations), thus largely ignoring some valuable skeleton patterns. 
To fill this gap, a few recent works \cite{rao2022skeleton,rao2021multi} construct skeleton graphs to model body-component relations in terms of structure and action. These methods typically require multi-stage \textit{non-parallel} relation modeling, $e.g.$, \cite{rao2021multi} models collaborative relations conditioned on the structural relations, while they cannot simultaneously mine different underlying relations. On the other hand, they usually leverage \textit{sequence-level} instances such as averaged features of sequential graphs \cite{rao2022skeleton} for representation learning,
which inherently limits their ability to exploit richer graph semantics from fine-grained ($e.g.,$ node-level) representations. For example, there usually exist strong correlations between nearby body-joint nodes within a local spatial-temporal graph context, which can \textit{prompt} learning unique and recognizable skeleton patterns for person re-ID. 

To address the above challenges, we propose a general Transformer-based Skeleton Graph prototype contrastive learning (TranSG) paradigm with structure-trajectory prompted reconstruction as shown in Fig. \ref{first}, which integrates different relational features of skeleton graphs and contrasts representative graph features to learn discriminative representations for person re-ID. Specifically, we first model 3D skeletons as graphs, and propose the \textit{Skeleton Graph Transformer (SGT)} to perform \textit{full-relation} learning of body-joint nodes, so as to simultaneously aggregate key relational features of body structure and motion into effective graph representations.
Second, a \textit{Graph Prototype Contrastive learning (GPC)} approach is proposed to contrast and learn the most representative skeleton graph features (defined as \textit{“graph prototypes”}) of each identity. By pulling together both \textit{sequence-level} and \textit{skeleton-level} graph representations to their corresponding prototypes while pushing apart representations of different prototypes, we encourage the model to capture discriminative graph features and high-level skeleton semantics ($e.g.,$ identity-associated patterns) for person re-ID. Last, motivated by the inherent structural correlations and pattern continuity of body joints \cite{rao2021self}, we devise a \textit{graph Structure-Trajectory Prompted Reconstruction (STPR) mechanism} to exploit the spatial-temporal context ($i.e.,$ graph structure and trajectory) of skeleton graphs to prompt the skeleton graph reconstruction, which facilitates learning richer graph semantics and more effective graph representations for the person re-ID task.

Our key contributions can be summarized as follows:
\begin{itemize}
    \item We present a generic TranSG paradigm to learn effective representations from skeleton graphs for person re-ID. To the best of our knowledge, TranSG is the first \textit{transformer} paradigm that unifies skeletal relation learning and skeleton graph contrastive learning specifically for skeleton-based person re-ID.
    \item We devise a skeleton graph transformer (SGT) to fully capture relations in skeleton graphs and integrate key correlative node features into graph representations.
    \item We propose the graph prototype contrastive learning (GPC) to contrast and learn the most representative graph features and identity-related semantics from both skeleton and sequence levels for person re-ID.
    \item We devise the graph structure-trajectory prompted reconstruction (STPR) to exploit spatial-temporal graph contexts for reconstruction, so as to capture more key graph semantics and unique features for person re-ID.
\end{itemize}

 Extensive experiments on five public benchmarks show that TranSG prominently outperforms existing state-of-the-art methods and is highly scalable to be applied to different graph modeling, RGB-estimated or unlabeled skeleton data.

\section{Related Works}
\textbf{Person Re-identification Using Skeleton Data.} 
Early methods manually design skeleton descriptors ($e.g.,$ anthropometric and gait attributes) from raw body joints. In \cite{barbosa2012re}, seven Euclidean distances between certain joints are computed to perform distance metric learning for person re-ID. An further extension with 13 ($D_{13}$) and 16 skeleton descriptors ($D_{16}$) in \cite{munaro2014one} and \cite{pala2019enhanced} are utilized to identify different persons. 
Recent methods employ deep neural networks to perform representation learning with skeleton sequences or graphs. In \cite{liao2020model}, a CNN-based model PoseGait is devised to encode body-joint sequences and hand-crafted pose features for human recognition. 
\cite{rao2020self} proposes the AGE model to encode recognizable gait features from 3D skeleton sequences, while its extension SGELA \cite{rao2021self} further combines self-supervised semantics learning ($e.g.,$ sequence sorting) and sequence contrastive learning to improve discriminative feature learning. The SimMC framework \cite{rao2022simmc} is proposed to encode prototypes and intra-sequence relations of masked skeleton sequences for person re-ID. MG-SCR \cite{rao2021multi} and SM-SGE \cite{rao2021sm} perform multi-stage body-component relation learning based on multi-scale graphs to learn person re-ID representations.
\cite{rao2022revisiting} proposes a general skeleton feature re-ranking mechanism for skeleton-based person re-ID.

\textbf{Contrastive Learning.}
 The aim of contrastive learning is to pull closer homogeneous or positive representation pairs while pushing farther negative pairs in a certain feature space. It has been broadly applied to various areas to learn effective feature representations \cite{wu2018unsupervised,li2021prototypical,chen2020a,he2020momentum,rao2021self,chen2021exploring,rao2022simmc,rao2022skeleton,rao2021augmented}. An instance discrimination paradigm with exemplar tasks is designed in \cite{wu2018unsupervised} for visual contrastive learning. A contrastive predictive coding (CPC) approach based on the context auto-encoding and probabilistic contrastive loss \cite{oord2018representation} is proposed to learn different-domain representations. 
PCL \cite{li2021prototypical} combines \textit{k}-means clustering and contrastive learning for image representation learning. Skeleton contrastive paradigms based on consecutive sequences or randomly masked sequences are devised in \cite{rao2021self} and \cite{rao2022simmc} for unsupervised skeleton representation learning and person re-ID.

\textbf{Prompt Learning.}
The function of prompts is to provide additional knowledge, instruction, or context for the input of models, such that they can be prompted to give more reliable outputs for different tasks \cite{petroni2019language,jia2021scaling,radford2021learning,zhou2022learning,jiang2020can,lester2021power,li2021prefix,liu2021pre,shin2020autoprompt}. 
In \cite{radford2021learning}, CLIP leverages language-based prompts to generalize the pre-trained visual representations to many tasks. CoOp \cite{zhou2022learning} is further devised to automatically model task-relevant prompts with continuous representations to improve the downstream task performance. 
As far as we know, this work for the first time explores \textit{graph prompts} (defined as structure and trajectory contexts) for skeleton graph reconstruction, so as to encourage capturing more key features and graph semantics ($e.g.,$ pattern continuity) for person re-ID. 

\begin{figure*}
    \centering
    \scalebox{0.472}{
    \includegraphics{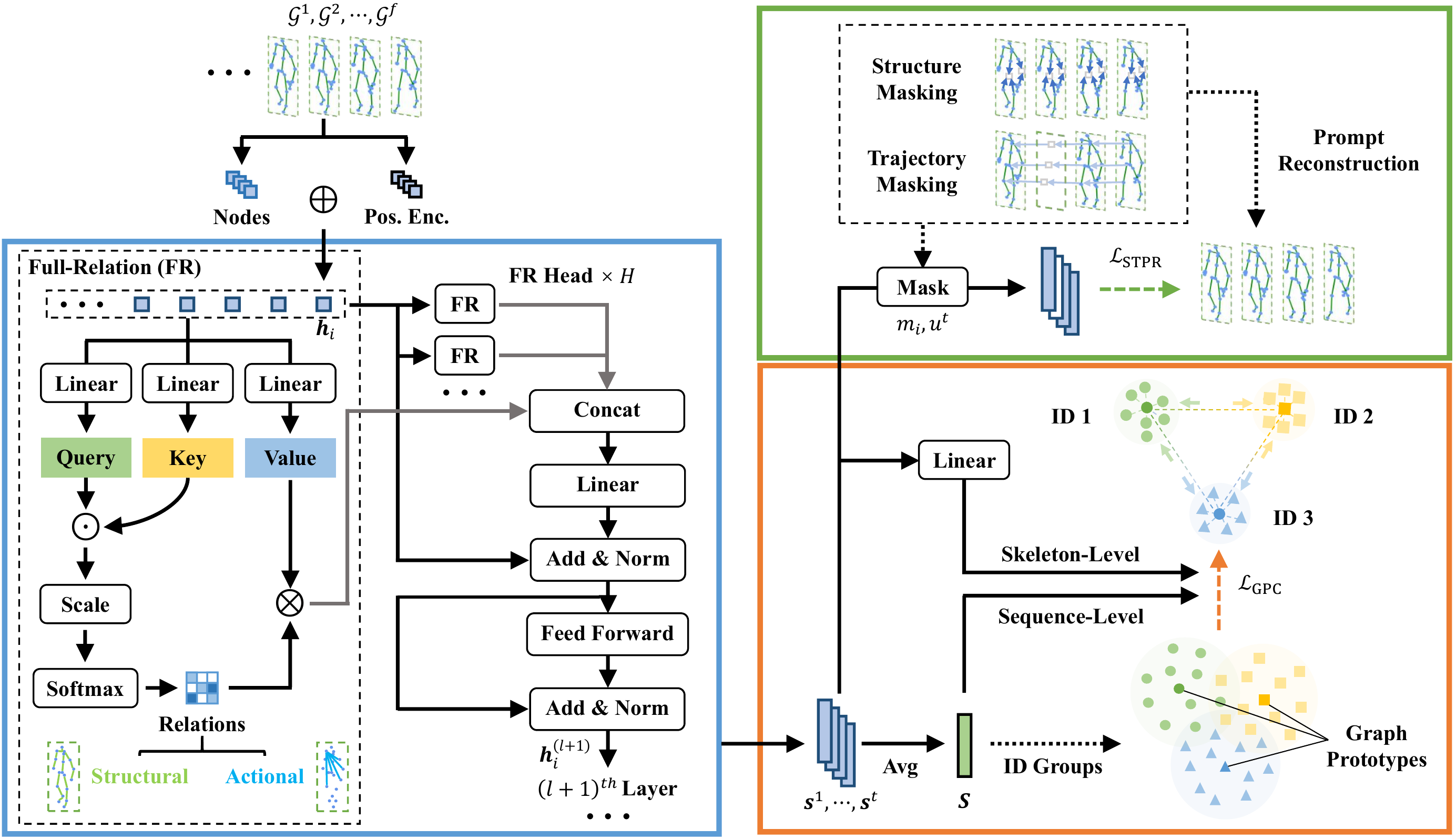}
    }
    \caption{Schematics of our approach with the skeleton graph transformer (SGT) (shown in the blue box), graph prototype contrastive learning (GPC) (presented in the orange box), and graph structure-trajectory prompted reconstruction (STPR) (presented in the green box). 
    }
    \label{model}
\end{figure*}

\section{The Proposed Approach}
Given a 3D skeleton sequence $\boldsymbol{X}\!=\!(\boldsymbol{x}_1,\cdots,\boldsymbol{x}_{f})\in \mathbb{R}^{f \times J \times 3}$, where $\boldsymbol{x}_{t}\in \mathbb{R}^{J \times 3}$ denotes the $t^{th}$ skeleton with 3D coordinates of $J$ body joints.
 Each skeleton sequence $\boldsymbol{X}$ corresponds to a person identity $\text{y}$, where $\text{y}\in \{1, \cdots, \text{C}\}$ and $\text{C}$ is the number of different classes ($i.e.$, identities). We use $\Phi_{T}=\left\{\boldsymbol{X}^{T}_{i}\right\}_{i=1}^{N_{1}}$, $\Phi_{P}=\left\{\boldsymbol{X}^{P}_{i}\right\}_{i=1}^{N_{2}}$, and $\Phi_{G}=\left\{\boldsymbol{X}^{G}_{i}\right\}_{i=1}^{N_{3}}$ to denote the \textit{training} set, \textit{probe} set, and \textit{gallery} set that contain $N_{1}$, $N_{2}$, and $N_{3}$ skeleton sequences of different persons in different scenes and views. Our aim is to learn an encoder to map skeleton sequences into effective representations, such that the encoded skeleton sequence representations (denoted as $\{\boldsymbol{S}^{P}_i\}_{i=1}^{N_{2}}$) in the probe set can be matched with the representations (denoted as $\{\boldsymbol{S}^{G}_i\}_{i=1}^{N_{3}}$) of the same identity in the gallery set.

The overview of our TranSG approach is shown in Fig. \ref{model}: Firstly, we represent each skeleton sequence $\boldsymbol{x}_1,\cdots,\boldsymbol{x}_{f}$ as skeleton graphs $\mathcal{G}^{1},\cdots,\mathcal{G}^{f}$ (see Sec. \ref{graph_construction}), and integrate the graph positional information into their node representations $\boldsymbol{h}_{i}$. Then, they are fed into the skeleton graph transformer (SGT) to fully capture body-component relations with multiple full-relation (FR) heads (see Sec. \ref{sec_SGT}). We employ multiple SGT layers and take the $l^{th}$ layer output $\boldsymbol{h}^{(l+1)}_{i}$ as the input of $(l+1)^{th}$ layer. Next, the centroids of graph features belonging to different identities (ID) are utilized to generate \textit{graph prototypes}, and we enhance the similarity of both skeleton-level ($\boldsymbol{s}^{t}$) and sequence-level graph representations ($\boldsymbol{S}$) to their corresponding prototypes, while maximizing their dissimilarity to other prototypes by optimizing $\mathcal{L}_{\mathrm{GPC}}$ (see Sec. \ref{sec_GPC}). Meanwhile, we randomly mask skeleton graph structure and node trajectory, and exploit correspondingly masked graph representations as contexts to prompt reconstruction by minimizing $\mathcal{L}_{\mathrm{STPR}}$ (see Sec. \ref{sec_ST_prompt}). The skeleton graph representations learned from our approach are exploited for person re-ID (see Sec. \ref{sec_entire}).

\subsection{Skeleton Graph Construction}
\label{graph_construction}
The human body with joints can be naturally modeled as graphs to characterize rich structural and positional information \cite{rao2021sm}. We construct skeleton graphs with the body joints as nodes and the structural connections between adjacent joints as edges. Each graph $\mathcal{G}^{t}(\mathcal{V}^{t}, \mathcal{E}^{t})$ (corresponding to the $t^{th}$ skeleton $\boldsymbol{x}_{t}$)  consists of nodes $\mathcal{V}^{t}=\{\boldsymbol{v}^{t}_{1}, \boldsymbol{v}^{t}_{2}, \cdots,\boldsymbol{v}^{t}_{J}\}$, $\boldsymbol{v}^{t}_{i}\in\mathbb{R}^{3}$, $i\in\{1,\cdots,J\}$ and edges $\mathcal{E}^{t}=\{e^{t}_{i,j}\ | \boldsymbol{v}^{t}_{i}, \boldsymbol{v}^{t}_{j}\!\in\!\mathcal{V}^{t}\}$, $e^{t}_{i,j}\in\mathbb{R}$. Here $\mathcal{V}^{t}$ and $\mathcal{E}^{t}$ denote the set of nodes corresponding to $J$ different body joints and the set of their internal connection relations, respectively. 
The adjacency matrix of $\mathcal{G}^{t}$ is denoted as $\mathbf{A}^{t} \in \mathbb{R}^{J \times J}$ to represent the relations among $J$ nodes. $\mathbf{A}^{t}$ is initialized based on the connections of adjacent body joints.

\subsection{Skeleton Graph Transformer}
\label{sec_SGT}
As our goal is to capture discriminative skeleton features for person re-ID, it is crucial to consider two \textit{unique} properties of human skeletons: (1) Body \textit{structural} features, which can be inferred from the relations between adjacent body joints; (2) Skeleton \textit{actional} patterns ($e.g.,$ gait \cite{murray1964walking}), which are typically characterized by the relations among different body components \cite{rao2021sm}. From the perspective of graphs, we regard each body-joint node as a basic body component, and propose to combine the above relation learning as a \textit{full-relation} learning of body-joint nodes, so as to fully aggregate key body and motion features from skeleton graphs.
For this purpose, we devise the \textit{skeleton graph transformer (SGT)} as follows (shown in Fig. \ref{model}).

Given a skeleton graph $\mathcal{G}^{t}(\mathcal{V}^{t}, \mathcal{E}^{t})$ and its adjacency matrix $\mathbf{A}^{t}$, we first exploit the pre-defined graph structure to generate the positional encoding for graph nodes with:
\begin{equation}
\Delta=\mathbf{I}-\mathbf{D}^{-1 / 2} \mathbf{A} \mathbf{D}^{-1 / 2}=\mathbf{U}^T \mathbf{\Lambda} \mathbf{U},
\label{eq_1}
\end{equation}
where $\mathbf{A}, \mathbf{D}$ are the adjacency matrix and degree matrix of the skeleton graph $\mathcal{G}$, respectively, and $\mathbf{\Lambda},\mathbf{U}$ denote the matrices of Laplacian eigenvalues and eigenvectors, respectively. $\mathbf{U}^T \mathbf{\Lambda} \mathbf{U}$ is the factorization of the graph Laplacian matrix. For convenience, we use $\mathbf{A}$ to represent $\mathbf{A}^{t}$ as skeleton graphs in the same dataset share the identical initialized adjacency matrix. We follow \cite{dwivedi2020generalization} to adopt the $K$ smallest non-trivial eigenvectors as the node positional encoding, denoted as $\boldsymbol{\lambda}_{i}\in\mathbb{R}^{K}$ for the node $\boldsymbol{v}_{i}$. They are mapped into feature spaces of the same dimension $d$ with the affine transformation, which are then added by:
\begin{equation}
\boldsymbol{h}_{i}=(\textbf{W}_{v}\boldsymbol{v}_{i}+\boldsymbol{b}_{v})+(\textbf{W}_{p}\boldsymbol{\lambda}_{i} + \boldsymbol{b}_{p}),
\label{eq_2}
\end{equation}
where $\boldsymbol{h}_{i}\in\mathbb{R}^{d}$ denotes the $i^{th}$ position-encoded node representation, $\textbf{W}_{v}\in \mathbb{R}^{d \times 3}, \textbf{W}_{p}\in \mathbb{R}^{d \times K}, \boldsymbol{b}_{v}, \boldsymbol{b}_{p}\in \mathbb{R}^{d}$ are the learnable parameters of the feature transformation for the $i^{th}$ node $\boldsymbol{v}_{i}$ and its corresponding positional encoding. 
Intuitively, the addition of positional encoding in Eq. (\ref{eq_2}) helps preserve the unique positional information of nodes based on the graph structure, $i.e.$, structurally nearby nodes are endowed with similar positional features while the farther nodes possess more dissimilar positional features, so as to encourage more effective node representation learning \cite{dwivedi2020generalization}.

Then, given the body-joint node representations, we capture their inherent relations using multiple \textit{independent} full-relation (FR) heads (see Fig. \ref{model}), and update node representations by aggregating corresponding relational features:
\begin{equation}
\boldsymbol{w}_{i, j}^{k,l}=\operatorname{Softmax}_j\left(\frac{(\boldsymbol{Q}^{k, l} \boldsymbol{h}_{i}^{(l)}) \cdot (\boldsymbol{K}^{k, l} \boldsymbol{h}_{j}^{(l)})}{\sqrt{d_\text{k}}}\right),
\label{eq_3}
\end{equation}
\begin{equation}
\boldsymbol{\hat{h}}_i^{(l)}=\boldsymbol{O}^{l} {\bigg\|}_{k=1}^{H}\left(\sum^{J}_{j=1} \boldsymbol{w}_{i,j}^{k, l} \boldsymbol{V}^{k, l} \boldsymbol{h}_{j}^{(l)}\right),
\label{eq_4}
\end{equation}
where $\boldsymbol{Q}^{k,l},\boldsymbol{K}^{k,l},\boldsymbol{V}^{k,l}\in\mathbb{R}^{d_{\text{k}}\times d}$ are the parameter matrices for query, key, and value transformations in the $k^{th}$ FR head of the $l^{th}$ SGT layer, $\boldsymbol{O}^{l} \in \mathbb{R}^{d \times d}$ is the parameter matrix for output transformation of the $l^{th}$ SGT layer.
$\frac{1}{\sqrt{d_{\text{k}}}}$ is the scaling factor for the scaled dot-product similarity, $\boldsymbol{w}_{i,j}^{k,l}$ denotes the normalized relation value between the $i^{th}$ and $j^{th}$ node captured by the $k^{th}$ FR head in the $l^{th}$ layer, 
${\big\|}$ represents the concatenation operation, and $H$ is the number of FR heads. For clarity, we use $\boldsymbol{\hat{h}}_i^{(l)}\in \mathbb{R}^{d}$ denotes the $i^{th}$ node representation that concatenates node features learned from different heads in the $l^{th}$ layer. 
It is worth noting that SGT naturally generalizes the \textit{self-attention} \cite{vaswani2017attention} to the full-relation learning of graph nodes, and can be viewed as a general paradigm that \textit{simultaneously} captures structural and actional relations from both adjacent and non-adjacent body-component nodes.
The multiple FR heads enable the model to \textit{jointly} attend to node relations from different feature subspaces and integrate more key correlative node features into final node representations.
We follow \cite{dwivedi2020generalization} to apply a Feed Forward Network (FFN) with residual connections \cite{he2016deep} and batch normalization \cite{ioffe2015batch} by:
\begin{equation}\boldsymbol{\overline{h}}_i^{(l)}=\operatorname{Norm}\left(\boldsymbol{h}_{i}^{(l)}+\boldsymbol{\hat{h}}_{i}^{(l)}\right),
\label{eq_5}
\end{equation}
\begin{equation}
\boldsymbol{h}_i^{(l+1)}=\operatorname{Norm}\left(\boldsymbol{\overline{h}}_i^{(l)}+\textbf{W}_{2}^{l} \ \sigma\left(\textbf{W}_{1}^{l} \boldsymbol{\overline{h}}_i^{(l)}\right)\right).
\label{eq_6}
\end{equation}
In Eq. (\ref{eq_5}) and (\ref{eq_6}), $\operatorname{Norm}(\cdot)$ denotes the batch normalization operation, $\textbf{W}_{1}^{l}\in \mathbb{R}^{2d \times d}, \textbf{W}_{2}^{l}\in \mathbb{R}^{d \times 2d}$ are the learnable parameters of FFN, $\sigma(\cdot)$ is the $\operatorname{ReLU}$ activation function, $\boldsymbol{\overline{h}}_i^{(l)}$ and $\boldsymbol{h}_i^{(l+1)}$ represent the intermediate and \textit{output} node representations of the $l^{th}$ SGT layer, respectively. We average the node features in each skeleton graph as the corresponding graph representation, and then integrate $f$ consecutive graph representations into the final sequence-level graph representation $\boldsymbol{S}$ with:
\begin{equation}
\boldsymbol{S}=\frac{1}{f}\sum^{f}_{t=1}\boldsymbol{s}^{t}=\frac{1}{f}\sum^{f}_{t=1}\frac{1}{J}\sum^{J}_{i=1} \boldsymbol{h}^{t}_{i},
\label{eq_7}
\end{equation}
where $\boldsymbol{S}, \boldsymbol{s}^{t} \in \mathbb{R}^{d}$ are the \textit{sequence-level} and \textit{skeleton-level} graph representations, corresponding to a skeleton sequence $\boldsymbol{X}$ and the $t^{th}$ skeleton $\boldsymbol{x}_{t}$ in the sequence, respectively. For simplicity of presentation, we use $\boldsymbol{h}^{t}_{i}$ to denote the encoded representation of $i^{th}$ node in the $t^{th}$ skeleton graph. Here we assume that each node representation with aggregated relational features (see Eq. (\ref{eq_3}), (\ref{eq_4})) contributes equally to the graph representation, and each skeleton graph has the same importance in representing patterns of an individual.

\subsection{Graph Prototype Contrastive Learning}
\label{sec_GPC}
Each individual's skeletons usually share the same anthropometric features ($e.g.,$ skeletal lengths), while their continuous sequence can characterize identity-specific walking patterns ($i.e.,$ gait) \cite{rao2022simmc}. In this context, it is desirable to mine the most representative skeleton features (defined as \textit{“prototypes”}) of each individual to learn distinguishable patterns. A straightforward way is to cluster sequence representations to mine prototypes for contrastive learning like \cite{rao2022skeleton,rao2022simmc}, while they can only generate identity-agnostic ($i.e.,$ pseudo-labeled) prototypes with large uncertainty or unreliability, $e.g.,$ when existing large intra-class variation, two same-class sequences with diverse patterns might be clustered to two prototype groups.
To encourage the model to generate representatives graph features more reliably, we propose to exploit the graph feature centroid of each \textit{ground-truth} identity as \textit{graph prototypes}, which are contrasted with both sequence-level and skeleton-level graph features to learn discriminative representations for person re-ID.

Given the encoded graph representations $\{\boldsymbol{S}^{T}_{i}\}_{i=1}^{N_{1}}$ of training skeleton sequences $\{\boldsymbol{X}^{T}_{i}\}_{i=1}^{N_{1}}$, we group them by ground-truth classes as $\{\widehat{\mathbb{S}}_{k}\}_{k=1}^{C}$, where $\widehat{\mathbb{S}}_{k}=\{\boldsymbol{S}_{k,j}\}_{j=1}^{n_k}$ denotes the set of graph representations belonging to the $k^{th}$ identity, $\boldsymbol{S}_{k,j}$ is the $j^{th}$ sequence-level graph representation, and $n_k$ represents the number of $k$-class sequences. Then, the graph prototype is generated by averaging the graph features of the same class with:
\begin{equation}
\boldsymbol{c}_{k}=\frac{1}{n_k}\sum^{n_k}_{j=1}\boldsymbol{S}_{k,j} \ ,
\label{eq_8}
\end{equation}
where $\boldsymbol{c}_{k}\in\mathbb{R}^{d}$ denotes the graph prototype of the $k^{th}$ identity. To focus on the representative graph prototype of each identity to learn discriminative identity-related semantics from both skeleton and sequence levels, we propose the graph prototype contrastive (GPC) loss as:
\begin{equation}
\mathcal{L}_{\mathrm{GPC}}=\alpha \mathcal{L}^{seq}_{\mathrm{GPC}} + (1-\alpha)\mathcal{L}^{ske}_{\mathrm{GPC}},
\label{eq_9}
\end{equation}
where
\begin{equation}
\scalebox{0.8}{
$
\begin{aligned}
\mathcal{L}^{seq}_{\mathrm{GPC}}=\frac{1}{N_{1}} \sum_{k=1}^{C} \sum_{j=1}^{n_k}-\log \frac{\exp \left(\boldsymbol{S}_{k, j} \cdot \boldsymbol{c}_{k} / \tau_{1}\right)}{\sum_{m=1}^{C} \exp \left(\boldsymbol{S}_{k, j} \cdot \boldsymbol{c}_{m} / \tau_{1}\right)},
\label{eq_10}
\end{aligned}
$
}
\end{equation}
\begin{equation}
\scalebox{0.8}{
$
\begin{aligned}
\mathcal{L}^{ske}_{\mathrm{GPC}}=\frac{1}{fN_{1}} \sum_{k=1}^{C} \sum_{j=1}^{n_k}\sum_{t=1}^{f}-\log \frac{\exp \left(\mathcal{F}_{1}\left(\boldsymbol{s}^{t}_{k, j}\right) \cdot \mathcal{F}_{2}\left(\boldsymbol{c}_{k}\right) / \tau_{2}\right)}{\sum_{m=1}^{C} \exp \left(\mathcal{F}_{1}\left(\boldsymbol{s}^{t}_{k, j}\right) \cdot \mathcal{F}_{2}\left(\boldsymbol{c}_{m}\right) / \tau_{2}\right)}.
\label{eq_11}
\end{aligned}
$
}
\end{equation}
In Eq. (\ref{eq_9}), $\alpha$ is the weight coefficient to fuse sequence-level ($\mathcal{L}^{seq}_{\mathrm{GPC}}$) and skeleton-level graph prototype contrastive learning ($\mathcal{L}^{ske}_{\mathrm{GPC}}$). In Eq. (\ref{eq_10}) and (\ref{eq_11}), $\boldsymbol{c}_{m}$ represents the $m$-class graph prototype, $\boldsymbol{s}^{t}_{k, j}$ denotes the graph representation of the $t^{th}$ skeleton in the sequence that corresponds to $\boldsymbol{S}_{k, j}$ belonging to the $k^{th}$ identity, $\tau_{1}, \tau_{2}$ are the temperatures for contrastive learning, and $\mathcal{F}_{1}(\cdot)$, $\mathcal{F}_{2}(\cdot)$ are linear projection heads to transform skeleton-level graph representations and graph prototypes into the same contrastive space $\mathbb{R}^{d}$. It should be noted that the graph prototypes are generated from higher level ($i.e.,$ sequence-level) representations and the learnable linear projection in Eq. (\ref{eq_11}) can be viewed as integrating related graph features from both levels for contrastive learning. 
The proposed GPC loss is essentially a generalized skeleton prototype contrastive loss that combines joint-level relation encoding (see Sec. \ref{sec_SGT}), skeleton-level and sequence-level graph learning (see Eq. (\ref{eq_9})). Its objective can be theoretically modeled as an Expectation-Maximization (EM) solution (see Appendix I).


\subsection{Graph Structure-Trajectory Prompted Reconstruction Mechanism}
\label{sec_ST_prompt}
To exploit more valuable graph features and high-level semantics ($e.g.,$ pattern consistency) from both spatial and temporal contexts of skeleton graphs, we propose a self-supervised \textit{graph Structure and Trajectory Prompted Reconstruction (STPR)} mechanism. Motivated by the structural correlations and local motion continuity \cite{rao2021self} of body components, we devise two graph context based prompts (see Fig. \ref{model}), namely (1) partial node positions of the \textit{same} graph and (2) partial node trajectory among \textit{continuous} graphs, to reconstruct the graph structure and dynamics.

\textbf{Graph Structure Prompted Reconstruction}. Given the $t^{th}$ skeleton graph representation with $J$ encoded nodes $(\boldsymbol{h}^{t}_{1},\cdots,\boldsymbol{h}^{t}_{J})$ encoded by SGT (see Eq. (\ref{eq_1})-(\ref{eq_6})), we first randomly mask node positions to generate the masked graph representation as:
\begin{equation}
\boldsymbol{\hat{s}}^{t}=\frac{1}{J-a}\sum^{J}_{i=1}{m}_{i}\boldsymbol{h}^{t}_{i},
\label{eq_12}
\end{equation}
where $a$ is the number of masks, $m_{i}$ is the binary mask value ($i.e.,$ $0$ for masking and $1$ for unmasking) applied on the $i^{th}$ node representation $\boldsymbol{h}^{t}_{i}$, and we have $\sum_{i=1}^{J}m_{i}=J-a$. With both spatial positional information (Eq. (\ref{eq_1}), (\ref{eq_2})) and relational features (Eq. (\ref{eq_3}), (\ref{eq_4})) integrated into each node, the masked graph representation $\boldsymbol{\hat{s}}^{t} \in \mathbb{R}^{d}$ in Eq. (\ref{eq_12}) retains the context of graph structure, which is then exploited to prompt the skeleton reconstruction by:
\begin{equation}
\boldsymbol{\hat{x}}^{t}=f_{s}(\boldsymbol{\hat{s}}^{t}),
\label{eq_13}
\end{equation}
where $f_{s}(\cdot)$ is an MLP network with one hidden layer for skeleton reconstruction with the structure prompt, and $\boldsymbol{\hat{x}}^{t}\in\mathbb{R}^{J\times3}$ is the predicted skeleton. It is worth noting that Eq. (\ref{eq_13}) not only utilizes the unmasked node representations to reconstruct their corresponding positions, but also exploits them as the context to predict the masked node positions. For conciseness, we denote the predicted $i^{th}$ training skeleton sequence as $\boldsymbol{\hat{X}}_{i}=\left(\boldsymbol{\hat{x}}^{1},\cdots,\boldsymbol{\hat{x}}^{f}\right) \in \mathbb{R}^{f\times J\times3}$.

\begin{table*}[t]
\centering
\caption{Performance comparison with state-of-the-art skeleton-based methods on different datasets. $\spadesuit$ denotes skeleton graph based methods, ${\dagger}$ indicates using hand-crafted descriptors, and ${\ddagger}$ refers to sequence representation learning models.
\textbf{Bold} denotes the best results. }
\label{formal_results}
\scalebox{0.6}{
\setlength{\tabcolsep}{2.13mm}{
\begin{tabular}{l|rrrr|rrrr|rrrr|rrrr|rrrr|rrrr}
\hline
\multirow{2}{*}{\textbf{Methods}}                             & \multicolumn{4}{c|}{\textbf{BIWI-S}}                                    & \multicolumn{4}{c|}{\textbf{BIWI-W}}                                    & \multicolumn{4}{c|}{\textbf{KS20}}                                      & \multicolumn{4}{c|}{\textbf{IAS-A}}                                     & \multicolumn{4}{c|}{\textbf{IAS-B}}                                     & \multicolumn{4}{c}{\textbf{KGBD}}                                       \\ \cline{2-25} 
                                                              & \textbf{mAP}  & \textbf{R$_{1}$} & \textbf{R$_{5}$} & \textbf{R$_{10}$} & \textbf{mAP}  & \textbf{R$_{1}$} & \textbf{R$_{5}$} & \textbf{R$_{10}$} & \textbf{mAP}  & \textbf{R$_{1}$} & \textbf{R$_{5}$} & \textbf{R$_{10}$} & \textbf{mAP}  & \textbf{R$_{1}$} & \textbf{R$_{5}$} & \textbf{R$_{10}$} & \textbf{mAP}  & \textbf{R$_{1}$} & \textbf{R$_{5}$} & \textbf{R$_{10}$} & \textbf{mAP}  & \textbf{R$_{1}$} & \textbf{R$_{5}$} & \textbf{R$_{10}$} \\ \hline
${D_{13}}^{\dagger}$ \cite{munaro2014one}    & 13.1          & 28.3             & 53.1             & 65.9              & 17.2          & 14.2             & 20.6             & 23.7              & 18.9          & 39.4             & 71.7             & 81.7              & 24.5          & 40.0             & 58.7             & 67.6              & 23.7          & 43.7             & 68.6             & 76.7              & 1.9           & 17.0             & 34.4             & 44.2              \\
${D_{16}}^{\dagger}$ \cite{pala2019enhanced} & 16.7          & 32.6             & 55.7             & 68.3              & 18.8          & 17.0             & 25.3             & 29.6              & 24.0          & 51.7             & 77.1             & 86.9              & 25.2          & 42.7             & 62.9             & 70.7              & 24.5          & 44.5             & 69.1             & 80.2              & 4.0           & 31.2             & 50.9             & 59.8              \\
PoseGait$^{\dagger}$\cite{liao2020model}     & 9.9           & 14.0             & 40.7             & 56.7              & 11.1          & 8.8              & 23.0             & 31.2              & 23.5          & 49.4             & 80.9             & \textbf{90.2}              & 17.5          & 28.4             & 55.7             & 69.2              & 20.8          & 28.9             & 51.6             & 62.9              & 13.9          & 50.6             & 67.0             & 72.6              \\
AGE$^{\ddagger}$\cite{rao2020self}           & 8.9           & 25.1             & 43.1             & 61.6              & 12.6          & 11.7             & 21.4             & 27.3              & 8.9           & 43.2             & 70.1             & 80.0              & 13.4          & 31.1             & 54.8             & 67.4              & 12.8          & 31.1             & 52.3             & 64.2              & 0.9           & 2.9              & 5.6              & 7.5               \\
SGELA$^{\ddagger}$\cite{rao2021self}         & 15.1          & 25.8             & 51.8             & 64.4              & 19.0          & 11.7             & 14.0             & 14.7              & 21.2          & 45.0             & 65.0             & 75.1              & 13.2          & 16.7             & 30.2             & 44.0              & 14.0          & 22.2             & 40.8             & 50.2              & 4.5           & 38.1             & 53.5             & 60.0              \\
MG-SCR$^{\spadesuit}$\cite{rao2021multi}     & 7.6           & 20.1             & 46.9             & 64.1              & 11.9          & 10.8             & 20.3             & 29.4              & 10.4          & 46.3             & 75.4             & 84.0              & 14.1          & 36.4             & 59.6             & 69.5              & 12.9          & 32.4             & 56.5             & 69.4              & 6.9           & 44.0             & 58.7             & 64.6              \\
SM-SGE$^{\spadesuit}$\cite{rao2021sm}        & 10.1          & 31.3             & 56.3             & 69.1              & 15.2          & 13.2             & 25.8             & 33.5              & 9.5           & 45.9             & 71.9             & 81.2              & 13.6          & 34.0             & 60.5             & 71.6              & 13.3          & 38.9             & 64.1             & 75.8              & 4.4           & 38.2             & 54.2             & 60.7              \\
SPC-MGR$^{\spadesuit}$\cite{rao2022skeleton} & 16.0          & 34.1             & 57.3             & 69.8              & 19.4          & 18.9             & 31.5             & 40.5              & 21.7          & 59.0             & 79.0             & 86.2              & 24.2          & 41.9             & 66.3             & 75.6              & 24.1          & 43.3             & 68.4             & 79.4              & 6.9           & 40.8             & 57.5             & 65.0              \\
SimMC$^{\ddagger}$\cite{rao2022simmc}        & 12.3          & 41.7             & 66.6             & 76.8              & 19.9          & 24.5             & 36.7             & 44.5              & 22.3          & 66.4             & 80.7             & 87.0              & 18.7          & 44.8             & 65.3             & 72.9              & 22.9          & 46.3             & 68.1             & 77.0              & 11.7          & 54.9             & 66.2             & 70.6              \\
\textbf{TranSG$^{\spadesuit}$ (Ours)}                         & \textbf{30.1} & \textbf{68.7}    & \textbf{86.5}    & \textbf{91.8}     & \textbf{26.9} & \textbf{32.7}    & \textbf{44.9}    & \textbf{52.2}     & \textbf{46.2} & \textbf{73.6}    & \textbf{86.3}    & \textbf{90.2}     & \textbf{32.8} & \textbf{49.2}    & \textbf{68.5}    & \textbf{76.2}     & \textbf{39.4} & \textbf{59.1}    & \textbf{77.0}    & \textbf{87.0}     & \textbf{20.2} & \textbf{59.0}    & \textbf{73.1}    & \textbf{78.2}     \\ \hline
\end{tabular}
}
}
\end{table*}

\textbf{Graph Trajectory Prompted Reconstruction}. To encourage the model to capture more unique temporal patterns from skeleton graphs, we propose to reconstruct graph trajectories based on their partial dynamics. In particular, we randomly mask the trajectory of each node with:
\begin{equation}
\boldsymbol{T}_{i}=\frac{1}{f-b}\sum^{f}_{t=1}{u}^{t}\boldsymbol{h}^{t}_{i},
\label{eq_14}
\end{equation}
where $\boldsymbol{T}_{i}\in\mathbb{R}^{d}$ is the \textit{masked} representation of $i$-node trajectory ($i.e.$, $\boldsymbol{h}^{1}_{i}, \cdots, \boldsymbol{h}^{f}_{i}$), $b$ is the number of trajectory masks, $u^{t}$ is the binary mask value applied to the $t^{th}$ position in the $i$-node trajectory ($i.e.$, $i^{th}$ node representation of the $t^{th}$ skeleton graph), and we have $\sum_{t=1}^{f}u^{t}=f-b$. The masked trajectory representation $\boldsymbol{T}_{i}$ preserves partial graph dynamics of different body-joint nodes, which are then used to prompt the skeleton reconstruction by: 
\begin{equation}
\boldsymbol{\overline{x}}_{i}=f_{t}(\boldsymbol{T}_{i}).
\label{eq_15}
\end{equation}
In Eq. (\ref{eq_15}), $\boldsymbol{\overline{x}}_{i} \in \mathbb{R}^{f\times3}$ denotes the temporal trajectory of the $i^{th}$ body joints predicted by an MLP network $f_{t}(\cdot)$ with one hidden layer. Based on the unmasked node representations over the temporal dimension, this reconstruction facilitates capturing key temporal dynamics and semantics ($e.g.,$ continuous patterns) to infer the node positions missed on the trajectory. We represent the predicted $i^{th}$ training sequence as $\boldsymbol{\overline{X}}_{i}\in \mathbb{R}^{f\times J\times3}$ by transposing the original predicted position matrix $\left(\boldsymbol{\overline{x}}_{1},\cdots,\boldsymbol{\overline{x}}_{J}\right) \in \mathbb{R}^{J\times f\times3}$.

We propose the STPR objective to combine both graph structure and trajectory prompted reconstruction as follows:
\begin{equation}
\mathcal{L}_{\mathrm{STPR}}=\beta\mathcal{L}_{\mathrm{STPR}}^{st}+(1-\beta)\mathcal{L}_{\mathrm{STPR}}^{tr},
\label{eq_16}
\end{equation}
where
\begin{equation}
\mathcal{L}_{\mathrm{STPR}}^{st}=\frac{1}{N_{1}}\sum_{i=1}^{N_{1}} \left\|\boldsymbol{\hat{X}}_{i}-\boldsymbol{X}^{T}_{i}\right\|_1,
\label{eq_17}
\end{equation}
\begin{equation}
\mathcal{L}_{\mathrm{STPR}}^{tr}
= \frac{1}{N_{1}}\sum_{i=1}^{N_{1}} \left\|\boldsymbol{\overline{X}}_{i}-\boldsymbol{X}^{T}_{i}\right\|_1.
\label{eq_18}
\end{equation}
In Eq. (\ref{eq_16}), $\beta$ is the weight coefficient to fuse structure-prompted ($\mathcal{L}_{\mathrm{STPR}}^{st}$) and trajectory-prompted reconstruction ($\mathcal{L}_{\mathrm{STPR}}^{tr}$). $\boldsymbol{X}^{T}_{i} \in \mathbb{R}^{f\times J \times 3}$ denotes the ground-truth positions of $i^{th}$ training skeleton sequence, and $\|\cdot\|_1$ represents the $\ell_1$ norm. We employ $\ell_1$ reconstruction loss following \cite{li2020dynamic,rao2021sm}, as it can alleviate gradient explosion of large losses while providing sufficient gradients for the positions with small losses to facilitate more precise reconstruction. 

\subsection{The Entire Approach}
\label{sec_entire}
We combine the proposed GPC (Sec. \ref{sec_GPC}) and STPR (Sec. \ref{sec_ST_prompt}) to perform skeleton representation learning with:
\begin{equation}
\mathcal{L}=\lambda \mathcal{L}_{\mathrm{GPC}}+(1-\lambda) \mathcal{L}_{\mathrm{STPR}} ,
\label{eq_19}
\end{equation}
where $\lambda$ is the weight coefficient to fuse two losses for training.
For person re-ID, we leverage the learned SGT to encode skeleton sequences of the probe set $\Phi_{P}$ into sequence-level graph representations, $\{\boldsymbol{S}^{P}_{i}\}_{i=1}^{N_{2}}$ (see Eq. (\ref{eq_7})), which are matched with the representations, $\{\boldsymbol{S}^{G}_{i}\}_{i=1}^{N_{3}}$, of the same identity in the gallery set $\Phi_{G}$ based on Euclidean distance.

\section{Experiments}
\label{experiments}
\subsection{Experimental Setups}
\textbf{Datasets.} Our approach is evaluated on four skeleton-based person re-ID benchmark datasets: \textit{IAS} \cite{munaro2014feature}, \textit{KS20} \cite{nambiar2017context}, \textit{BIWI} \cite{munaro2014one}, \textit{KGBD} \cite{andersson2015person}, which contain 11, 20, 50, and 164 different persons. We also verify the generality of TranSG on the RGB-estimated skeleton data from a large-scale multi-view gait dataset \textit{CASIA-B} \cite{yu2006framework} with 124 individuals under three conditions (Normal (N), Bags (B), Clothes (C)). We adopt the commonly-used probe and gallery settings following \cite{rao2022simmc,liu2015enhancing} for a fair comparison. 

\textbf{Implementation Details.} The number of body joints is $J=20$ in KGBD, IAS, BIWI, $J=25$ in KS20, and $J=14$ in the estimated skeleton data of CASIA-B.
The sequence length is set to $f=6$ for the Kinect-based skeleton datasets (IAS, KS20, BIWI, and KGBD) and $f=40$ for the RGB-estimated skeleton data (CASIA-B), following existing methods for a fair comparison. The embedding size of each node representation is $d=128$. We empirically employ 2 SGT layers with $H=8$ FR heads and $d_{\text{k}}=16$ for each layer. Each component in our approach is equally fused with $\alpha=\beta=\lambda=0.5$. For models trained with RGB-estimated skeletons, we set $\alpha=1.0$. $\tau_{1}=0.07$ and $\tau_{2}=14$ are empirically used for contrastive learning. The numbers of random masks are set to $a=10$ and $b=2$. We use the Adam optimizer with a learning rate of $3.5\times 10^{-4}$ for model optimization, and the batch size is set to 256 for all datasets. More details are provided in Appendix II.

\textbf{Evaluation Metrics.}
We compute the Cumulative Matching Characteristics (CMC) curve and adopt Rank-1 accuracy (R$_{1}$), Rank-5 accuracy (R$_{5}$), and Rank-10 accuracy (R$_{10}$) as performance metrics.  Mean Average Precision (mAP) \cite{zheng2015scalable} is also used to evaluate the overall performance.

\subsection{Comparison with State-of-the Art Methods}

We compare our approach with state-of-the-art graph-based methods, hand-crafted methods, and sequence learning methods on BIWI, KS20, IAS, and KGBD in Table \ref{formal_results}.


\textbf{Comparison with Graph-based Methods:}
As shown in Table \ref{formal_results}, the proposed TranSG significantly outperforms two state-of-the-art graph-based methods, MG-SCR \cite{rao2021multi} and SM-SGE \cite{rao2021sm}, by $11.7$-$36.7\%$ for mAP and $12.8$-$48.6\%$ for Rank-1 accuracy on different datasets. As these methods rely on multi-stage relation modeling and sequence-level context learning, the results demonstrate that the proposed full-relation learning model (SGT) with fine-grained ($i.e.,$ graph and node level) semantics learning (STPR) is able to learn more unique skeleton features for person re-ID. Compared with the latest SPC-MGR model \cite{rao2022skeleton} that utilizes sequence-level graph representations for contrastive learning, our model achieves superior performance with a large margin of $7.5\%$ to $24.5\%$ for mAP and $7.3\%$ to $34.6\%$ for Rank-1 accuracy on all datasets.
This suggests the higher effectiveness of our approach combining both sequence-level and skeleton-level graph prototype contrastive learning. We will also show the generality of our model under different-scale skeleton graph modeling in Sec. \ref{further_analysis}.

\textbf{Comparison with Hand-crafted and Sequence Learning Methods:}
In contrast to the methods using hand-crafted anthropometric descriptors ($D_{13}$ \cite{munaro2014one}, $D_{16}$ \cite{pala2019enhanced}) or 3D pose features (PoseGait \cite{liao2020model}), our approach consistently achieves higher performance by up to $27.3\%$ for mAP and $54.7\%$ for Rank-1 accuracy on all datasets.
TranSG also achieves a remarkable improvement over existing sequence representation learning models (AGE \cite{rao2020self}, SGELA \cite{rao2021self}, SimMC \cite{rao2022simmc}) in terms of mAP ($7.0$-$37.3\%$), Rank-1 ($4.1$-$56.1\%$), Rank-5 ($3.2$-$67.5\%$), and Rank-10 accuracy ($3.2$-$70.7\%$). This demonstrates the superiority of our graph-based model, as it can fully capture body relations and discriminative patterns from skeleton graphs for the person re-ID task.


\subsection{Ablation Study}
\begin{table}[t]
\centering
\caption{Ablation study with different configurations: Na\"ive prototype contrastive learning (PC), skeleton graph transformer (SGT) with direct supervised learning (DS) or graph prototype contrastive learning (GPC), and structure-trajectory prompted reconstruction (STPR). “+” indicates employing the component. }
\label{ablation_results}
\scalebox{0.6}{
\setlength{\tabcolsep}{2.85mm}{
\begin{tabular}{lrrrrrrrr}
\hline
\multirow{2}{*}{\textbf{Configurations}} & \multicolumn{2}{c}{\textbf{BIWI-S}} & \multicolumn{2}{c}{\textbf{BIWI-W}} & \multicolumn{2}{c}{\textbf{KS20}} & \multicolumn{2}{c}{\textbf{KGBD}} \\
                                         & \textbf{mAP}   & \textbf{R$_{1}$}   & \textbf{mAP}   & \textbf{R$_{1}$}   & \textbf{mAP}  & \textbf{R$_{1}$}  & \textbf{mAP}  & \textbf{R$_{1}$}  \\ \hline
\textbf{Baseline}                        & 9.3            & 24.8               & 14.1           & 10.9               & 9.5           & 17.0              & 6.4           & 34.5              \\
\textbf{PC}                              & 11.3           & 38.1               & 18.3           & 21.2               & 20.5          & 64.8              & 11.0          & 53.0              \\
\textbf{SGT + DS}                        & 19.0           & 42.4               & 21.1           & 21.7               & 27.6          & 60.0              & 11.1          & 51.5              \\
\textbf{SGT + GPC}                       & 26.7           & 66.6               & 25.5           & 31.2               & 42.5          & 71.3              & 18.1          & 57.0              \\
\textbf{SGT + GPC + STPR}                & 30.1           & 68.7               & 26.9           & 32.7               & 46.2          & 73.6              & 20.2          & 59.0              \\ \hline
\end{tabular}
}
}
\end{table}
We conduct ablation study to verify the effectiveness of each component in our approach. The skeleton sequences of concatenated joints are adopted as the baseline, and we include the na\"ive prototype contrastive learning (PC) using original sequences \cite{rao2022simmc} for comparison. As shown in Table \ref{ablation_results}, compared with using raw skeleton sequences or PC without graph modeling, applying SGT achieves significantly better performance in most cases, regardless of using GPC or not.
This demonstrates the effectiveness of the skeleton graph learning with SGT, as it can fully capture relations within skeletons to learn unique body structure and motion features for person re-ID. 
The SGT employing GPC achieves superior results than “SGT + DS” that uses direct supervised learning ($i.e.,$ cross-entropy loss) with an improvement of $4.4$-$14.9\%$ for mAP and $5.5$-$24.2\%$ for Rank-1 accuracy, which verifies the key role of the graph prototype contrastive learning in capturing more representative discriminative graph features from different identities. Finally, adding STPR consistently boosts model performance by $1.4$-$3.7\%$ for mAP on all dataset, which further demonstrates its effectiveness on capturing more valuable graph semantics and discriminative patterns for person re-ID.

\section{Further Analysis}
\label{further_analysis}
\begin{table}[t]
\centering
\caption{Performance comparison with \textit{appearance} or \textit{skeleton} based methods on CASIA-B. ``B-N'' denotes matching the ``Bags (B)'' probe set with the``Normal (N)'' gallery set. ``—'' indicates no published result. Full results are provided in Appendix II. }
\label{CASIA_B_results}
\scalebox{0.6}{
\setlength{\tabcolsep}{1.45mm}{
\begin{tabular}{ll|rrrrrrrrrr}
\hline
\textbf{}                                                           & \textbf{Probe-Gallery}                                     & \multicolumn{2}{c}{\textbf{N-N}} & \multicolumn{2}{c}{\textbf{B-B}} & \multicolumn{2}{c}{\textbf{C-C}} & \multicolumn{2}{c}{\textbf{C-N}} & \multicolumn{2}{c}{\textbf{B-N}} \\ \hline
\textbf{}                                                           & \textbf{Methods}                                           & \textbf{mAP}       & \textbf{R$_{1}$}      & \textbf{mAP}     & \textbf{R$_{1}$}    & \textbf{mAP}        & \textbf{R$_{1}$}       & \textbf{mAP}       & \textbf{R$_{1}$}       & \textbf{mAP}      & \textbf{R$_{1}$}     \\ \hline
\multicolumn{1}{l|}{\multirow{6}{*}{\rotatebox{90}{\textit{\textbf{Appearance}}}}} & LMNN \cite{weinberger2009distance}        & —                  & 3.9                   & —                & 18.3                & —                   & 17.4                   & —                  & 11.6                   & —                 & 23.1                 \\
\multicolumn{1}{l|}{}                                               & ITML \cite{davis2007information}          & —                  & 7.5                   & —                & 19.5                & —                   & 20.1                   & —                  & 10.3                   & —                 & 21.8                 \\
\multicolumn{1}{l|}{}                                               & ELF \cite{gray2008viewpoint}              & —                  & 12.3                  & —                & 5.8                 & —                   & 19.9                   & —                  & 5.6                    & —                 & 17.1                 \\
\multicolumn{1}{l|}{}                                               & SDALF \cite{farenzena2010person}          & —                  & 4.9                   & —                & 10.2                & —                   & 16.7                   & —                  & 11.6                   & —                 & 22.9                 \\
\multicolumn{1}{l|}{}                                               & MLR (Scores) \cite{liu2015enhancing}   & —                  & 13.6                  & —                & 13.6                & —                   & 13.5                   & —                  & 9.7                    & —                 & 14.7                 \\
\multicolumn{1}{l|}{}                                               & MLR (Features) \cite{liu2015enhancing} & —                  & 16.3                  & —                & 18.9                & —                   & 25.4                   & —                  & 20.3                   & —                 & 31.8                 \\ \hline
\multicolumn{1}{l|}{\multirow{6}{*}{\rotatebox{90}{\textit{\textbf{Skeleton}}}}}   & AGE \cite{rao2020self}                    & 3.5                & 20.8                  & 9.8              & 37.1                & 9.6                 & 35.5                   & 3.0                & 14.6                   & 3.9               & 32.4                 \\
\multicolumn{1}{l|}{}                                               & SM-SGE \cite{rao2021sm}                   & 6.6                & 50.2                  & 9.3              & 26.6                & 9.7                 & 27.2                   & 3.0                & 10.6                   & 3.5               & 16.6                 \\
\multicolumn{1}{l|}{}                                               & SPC-MGR \cite{rao2022skeleton}            & 9.1                & 71.2                  & 11.4             & 44.3                & 11.8                & 48.3                   & 4.3                & 22.4                   & 4.6               & 28.9                 \\
\multicolumn{1}{l|}{}                                               & SGELA \cite{rao2021self}                  & 9.8                & 71.8                  & 16.5             & 48.1                & 7.1                 & 51.2                   & 4.7                & 15.9                   & 6.7               & 36.4                 \\
\multicolumn{1}{l|}{}                                               & SimMC \cite{rao2022simmc}                 & 10.8               & \textbf{84.8}                  & 16.5             & \textbf{69.1}       & \textbf{15.7}                & \textbf{68.0}                   & 5.4                & \textbf{25.6}                   & 7.1               & 42.0                 \\
\multicolumn{1}{l|}{}                                               & TranSG (Ours)                                              & \textbf{13.1}      & 78.5         & \textbf{17.9}    & 67.1                & \textbf{15.7}       & 65.6          & \textbf{6.7}       & 23.0          & \textbf{8.6}      & \textbf{44.1}        \\ \hline
\end{tabular}
}
}
\end{table}

\textbf{Evaluation on RGB-estimated Skeletons.}
To verify the generality of our skeleton-based model on RGB-estimated skeletons, we utilize pre-trained pose estimation models to extract skeleton data from RGB videos of CASIA-B \cite{liao2020model}, and evaluate the performance of TranSG under different conditions. As presented in Table \ref{CASIA_B_results}, our approach not only outperforms many existing state-of-the-art skeleton-based models with a prominent improvement in most conditions, but also achieves superior performance to representative classical appearance-based methods that utilize RGB-based features ($e.g.,$ textures, silhouettes) or/and visual metric learning \cite{weinberger2009distance,davis2007information,gray2008viewpoint,farenzena2010person,liu2015enhancing}. 
This demonstrates the stronger ability of TranSG on capturing more discriminative features from the estimated skeletons, and also suggests its great potential to be applied to more general RGB-based scenarios.

\textbf{Application to Skeleton Graphs with Varying Scales.} 
To validate the effectiveness of our approach under different graph modeling, we follow \cite{rao2021sm} to construct different-scale graphs for model learning. As shown in Table \ref{topology_results}, TranSG significantly outperforms the state-of-the-art framework SM-SGE \cite{rao2021sm} when utilizing the original skeleton graphs (corresponding to $J$ joints) or higher level graphs with less nodes.
 This demonstrates that our model is compatible with different-level skeletal structures and can learn more effective features even under different graph modeling.

\begin{table}[t]
\centering
\caption{Performance of our approach on different-scale skeleton graphs. $J$, 10 or 5 nodes correspond to the joint-scale, part-scale or body-scale skeletal structures used in SM-SGE \cite{rao2021sm}.}
\label{topology_results}
\scalebox{0.6}{
\setlength{\tabcolsep}{2.45mm}{
\begin{tabular}{clrrrrrrrr}
\hline
\multirow{2}{*}{\textbf{Nodes}} & \multirow{2}{*}{\textbf{Methods}} & \multicolumn{2}{c}{\textbf{BIWI-S}} & \multicolumn{2}{c}{\textbf{BIWI-W}} & \multicolumn{2}{c}{\textbf{KS20}} & \multicolumn{2}{c}{\textbf{KGBD}} \\
                                &                                   & \textbf{mAP}    & \textbf{R$_{1}$}  & \textbf{mAP}    & \textbf{R$_{1}$}  & \textbf{mAP}   & \textbf{R$_{1}$} & \textbf{mAP}   & \textbf{R$_{1}$} \\ \hline
\multirow{2}{*}{\textbf{$J$}}   & SM-SGE \cite{rao2021sm}                            & 10.0            & 33.0              & 14.9            & 12.9              & 10.2           & 44.7             & 4.3            & 40.2             \\
                                & \textbf{TranSG (Ours)}            & \textbf{30.1}   & \textbf{68.7}     & \textbf{26.9}   & \textbf{32.7}     & \textbf{46.2}  & \textbf{73.6}    & \textbf{20.2}  & \textbf{59.0}    \\ \hline
\multirow{2}{*}{\textbf{10}}    & SM-SGE                            & 11.1            & 32.8              & 16.5            & 14.5              & 9.8            & 43.2             & 4.1            & 33.0             \\
                                & \textbf{TranSG (Ours)}            & \textbf{15.1}   & \textbf{37.9}     & \textbf{18.7}   & \textbf{20.0}     & \textbf{17.0}   & \textbf{48.1}    & \textbf{4.9}   & \textbf{36.9}    \\ \hline
\multirow{2}{*}{\textbf{5}}     & SM-SGE                            & 10.0            & 27.5              & 13.8            & 12.6              & 9.3            & 37.3             & 4.4            & 31.5             \\
                                & \textbf{TranSG (Ours)}            & \textbf{13.5}   & \textbf{36.6}     & \textbf{14.1}   & \textbf{17.0}     & \textbf{13.2}  & \textbf{40.8}    & \textbf{4.6}   & \textbf{34.6}    \\ \hline
\end{tabular}
}
}
\end{table}

\begin{table}[t]
\centering
\caption{Performance of our approach using only unlabeled skeletons. We use the same clustering setting in \cite{rao2022skeleton} for comparison.}
\label{unsupervised_results}
\scalebox{0.6}{
\setlength{\tabcolsep}{3.25mm}{
\begin{tabular}{lrrrrrrrr}
\hline
\multirow{2}{*}{\textbf{Methods}} & \multicolumn{2}{c}{\textbf{BIWI-S}} & \multicolumn{2}{c}{\textbf{BIWI-W}} & \multicolumn{2}{c}{\textbf{KS20}} & \multicolumn{2}{c}{\textbf{KGBD}} \\
                                  & \textbf{mAP}    & \textbf{R$_{1}$}  & \textbf{mAP}    & \textbf{R$_{1}$}  & \textbf{mAP}   & \textbf{R$_{1}$} & \textbf{mAP}  & \textbf{R$_{1}$}  \\ \hline
SPC-MGR \cite{rao2022skeleton}                          & \textbf{16.0}   & 34.1              & 19.4            & 18.9              & 21.7           & 59.0             & 6.9           & 40.8              \\
TranSG (Ours)                     & 14.2            & \textbf{42.2}     & \textbf{21.9}   & \textbf{26.6}     & \textbf{23.6}  & \textbf{65.0}    & \textbf{7.5}  & \textbf{43.6}     \\ \hline
\end{tabular}
}
}
\end{table}

\textbf{Transfer to Unsupervised Scenarios.} To apply our approach in an unsupervised manner without using ground-truth labels, we follow \cite{rao2022skeleton} to perform DBSCAN clustering \cite{ester1996density} of graph representations, and leverage their pseudo classes to generate graph prototypes for contrastive learning. With only \textit{unlabeled} skeletons as inputs, TranSG can still achieve superior performance than the latest graph-based method SPC-MGR \cite{rao2022skeleton} in most cases, as shown in Table \ref{unsupervised_results}. 
This further demonstrates the generality of our approach, which could be promisingly transferred to more related tasks such as unsupervised open-set person re-ID.


 \begin{figure}[t]
    \centering
    \scalebox{0.15}{\includegraphics[]{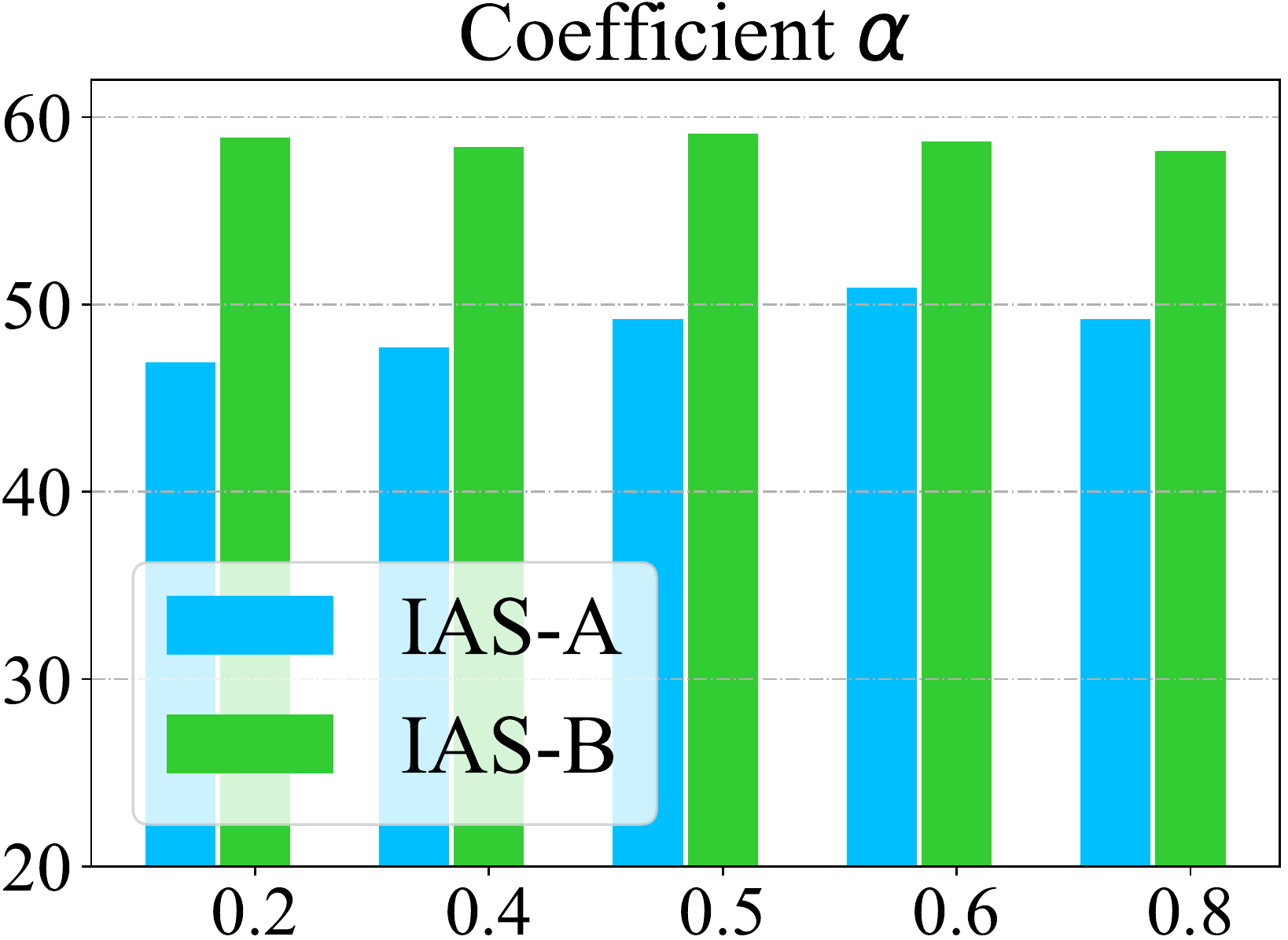}} \ \
  \scalebox{0.15}{\includegraphics[]{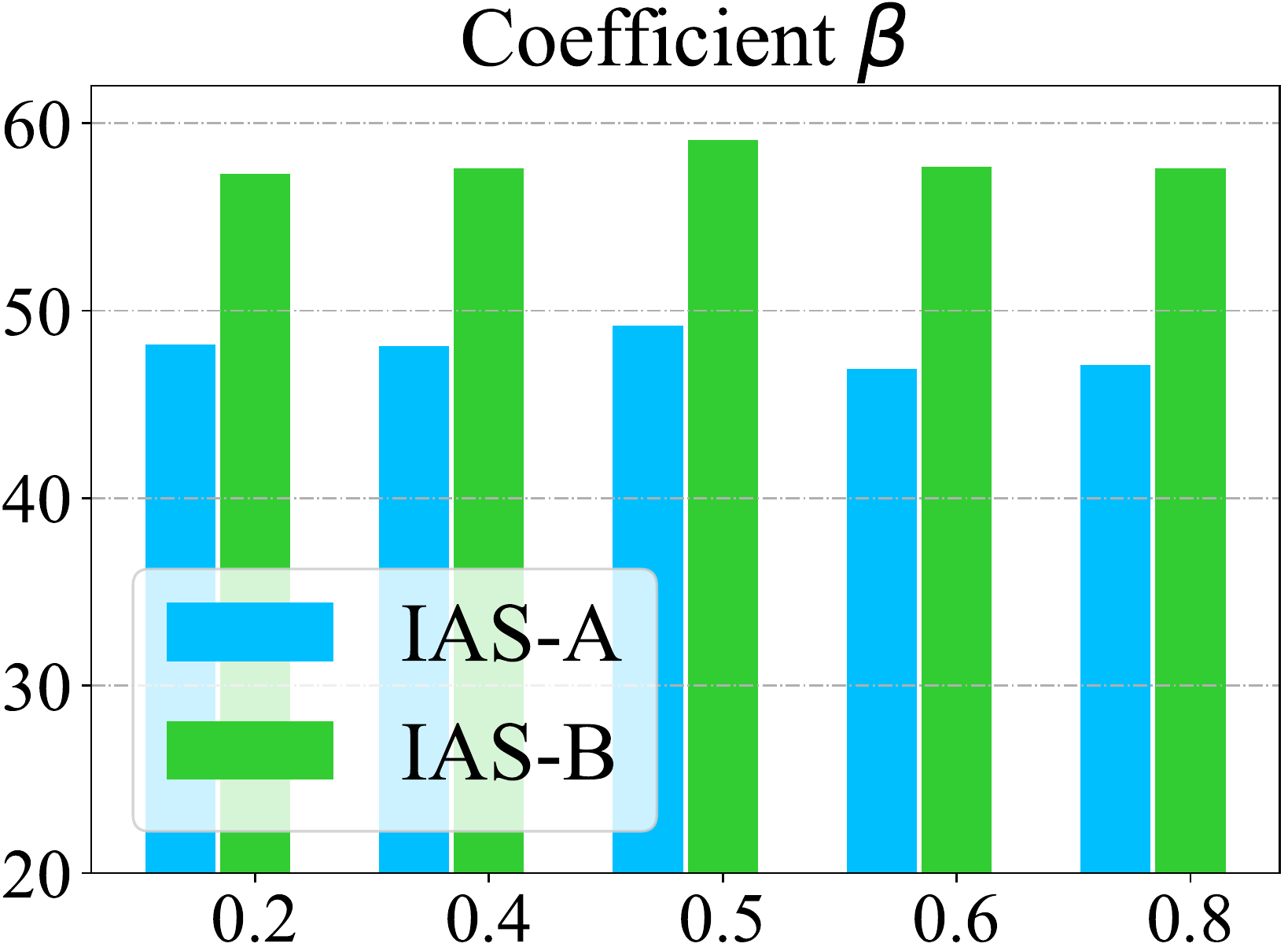}} \ \
  \scalebox{0.15}{\includegraphics[]{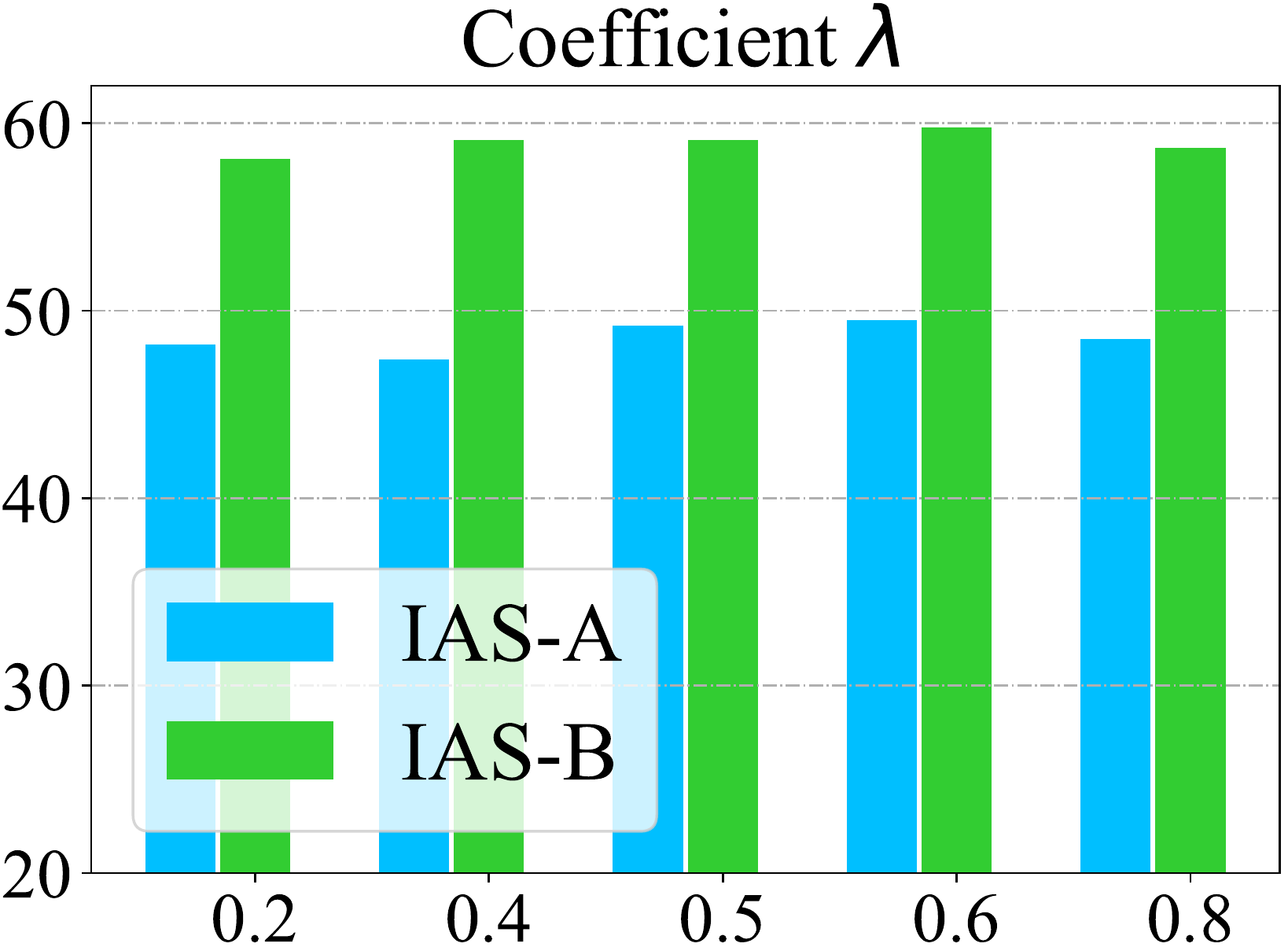}} \ \
      \scalebox{0.15}{\includegraphics[]{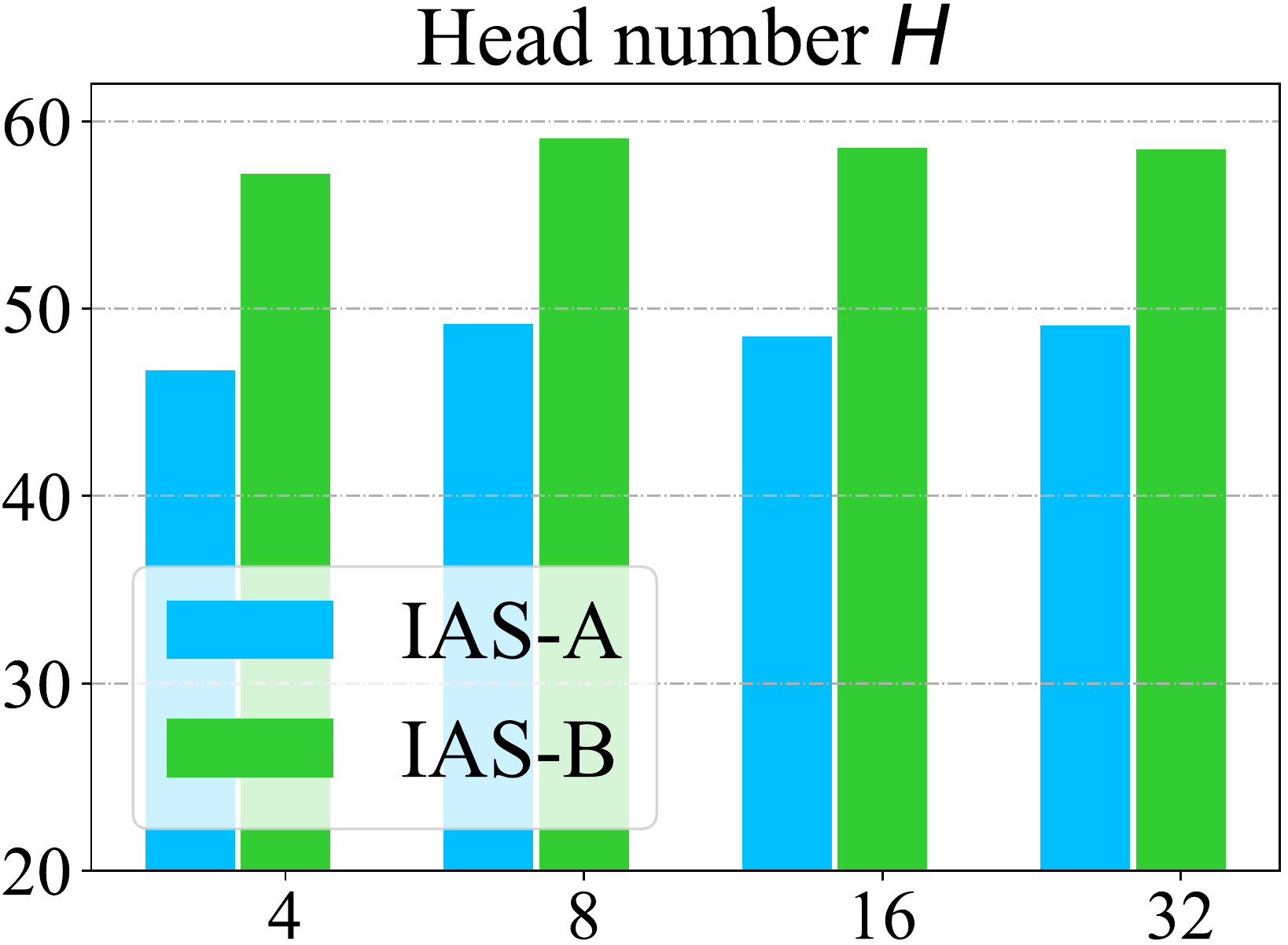}} \ \ 
         \scalebox{0.15}{\includegraphics[]{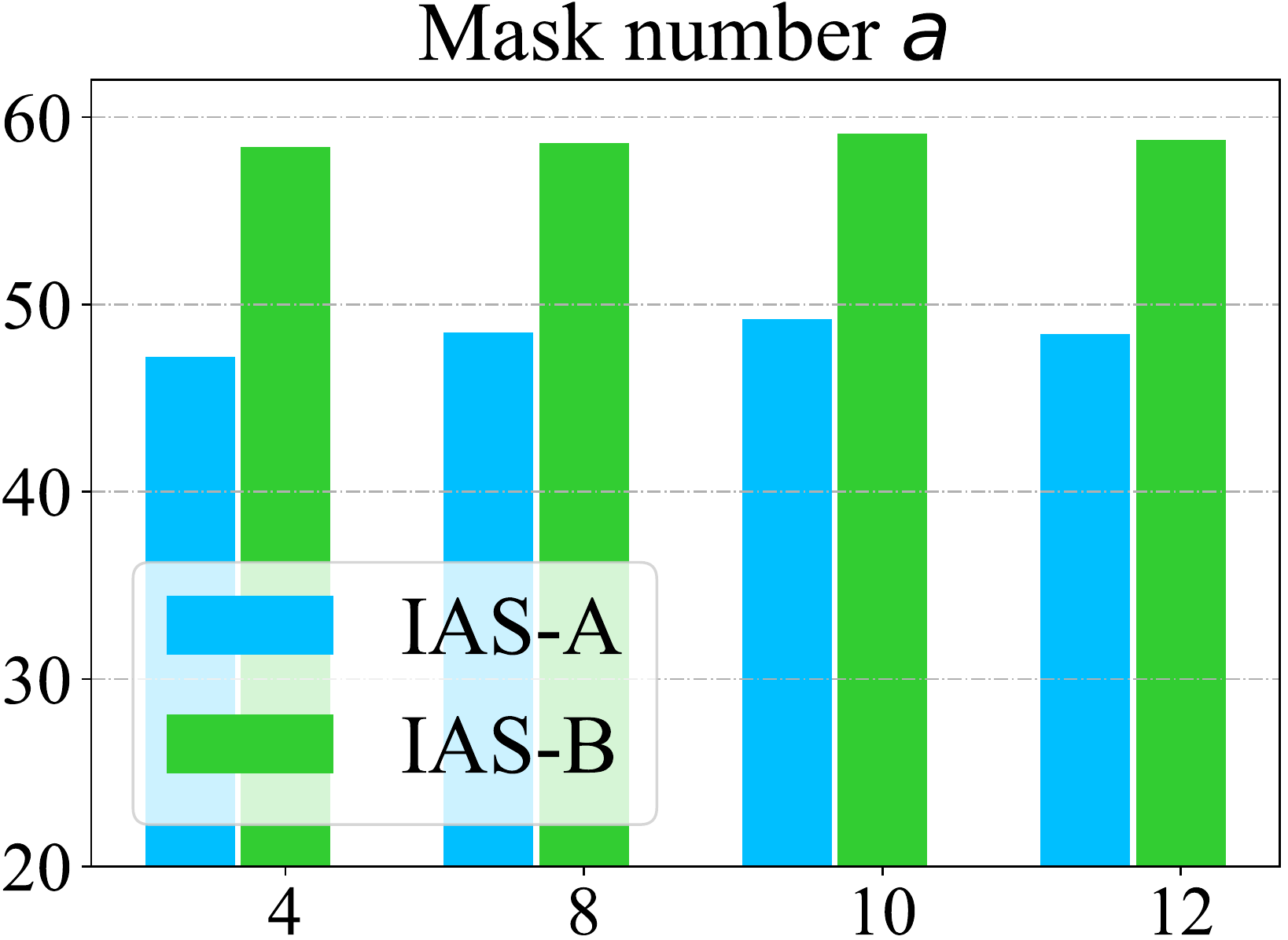}} \ \
         \scalebox{0.15}{\includegraphics[]{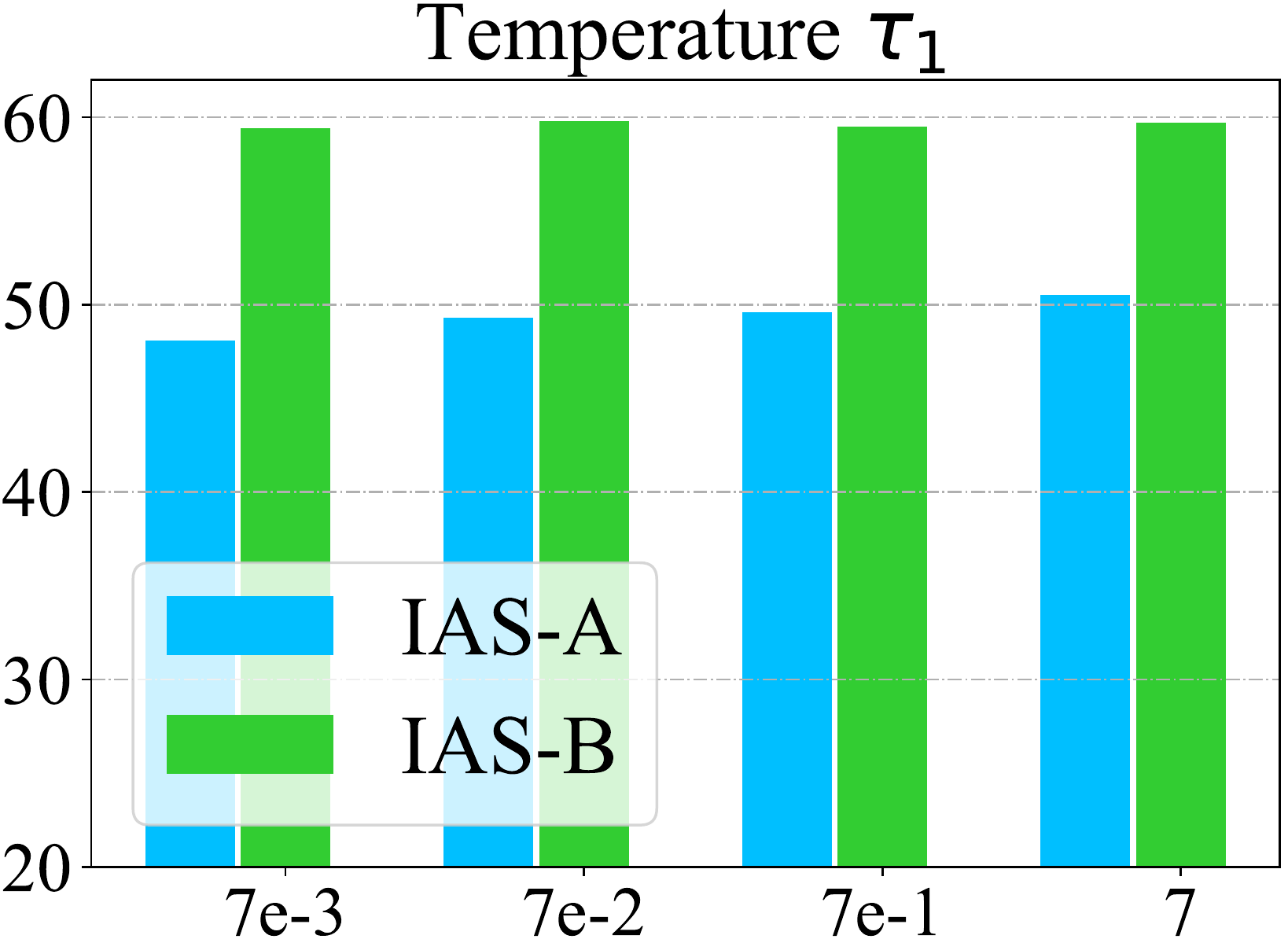}}
    \caption{Rank-1 accuracy of our approach on different probe sets (IAS-A and IAS-B) when setting different hyper-parameters.}
    \label{parameters}
\end{figure}

\begin{figure}
\centering
\subcaptionbox{SPC-MGR \cite{rao2022skeleton}}{\scalebox{0.17}{\includegraphics[]{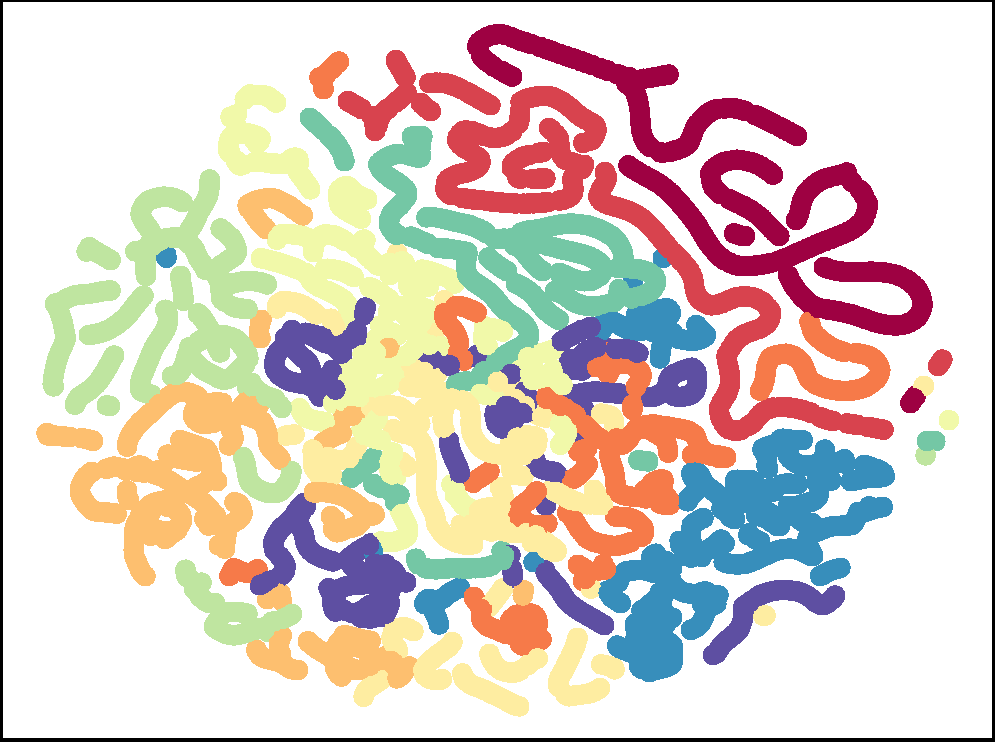}}}
 \quad
	\subcaptionbox{SimMC \cite{rao2022simmc}}{\scalebox{0.17}{\includegraphics[]{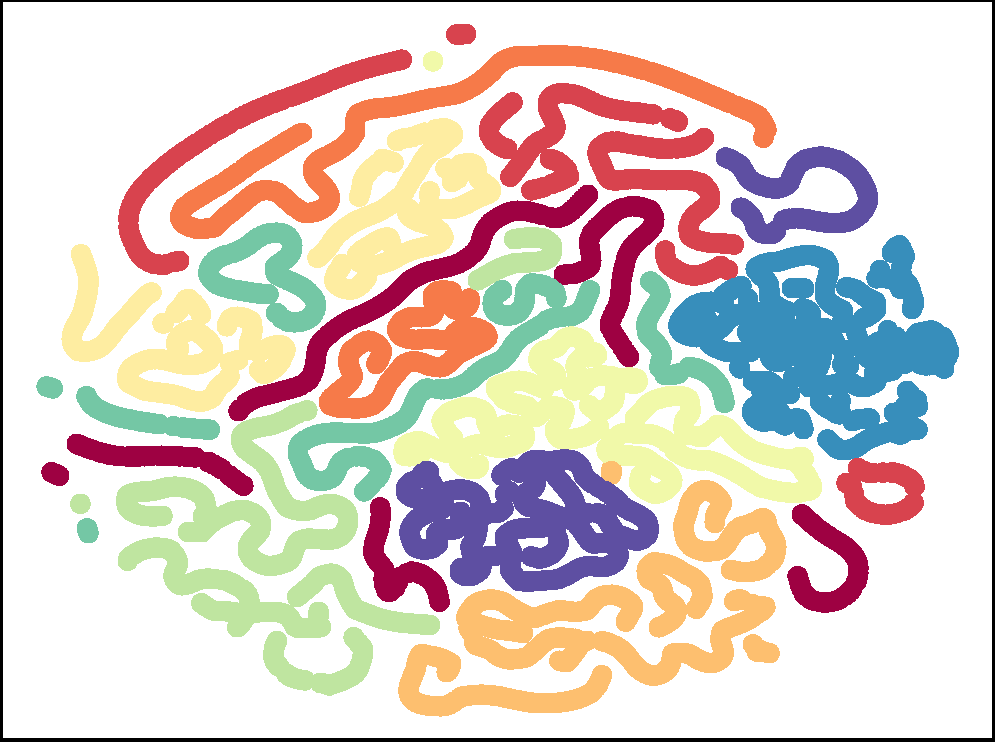}}}
    \quad
	\subcaptionbox{TranSG (Ours)}{\scalebox{0.17}{\includegraphics[]{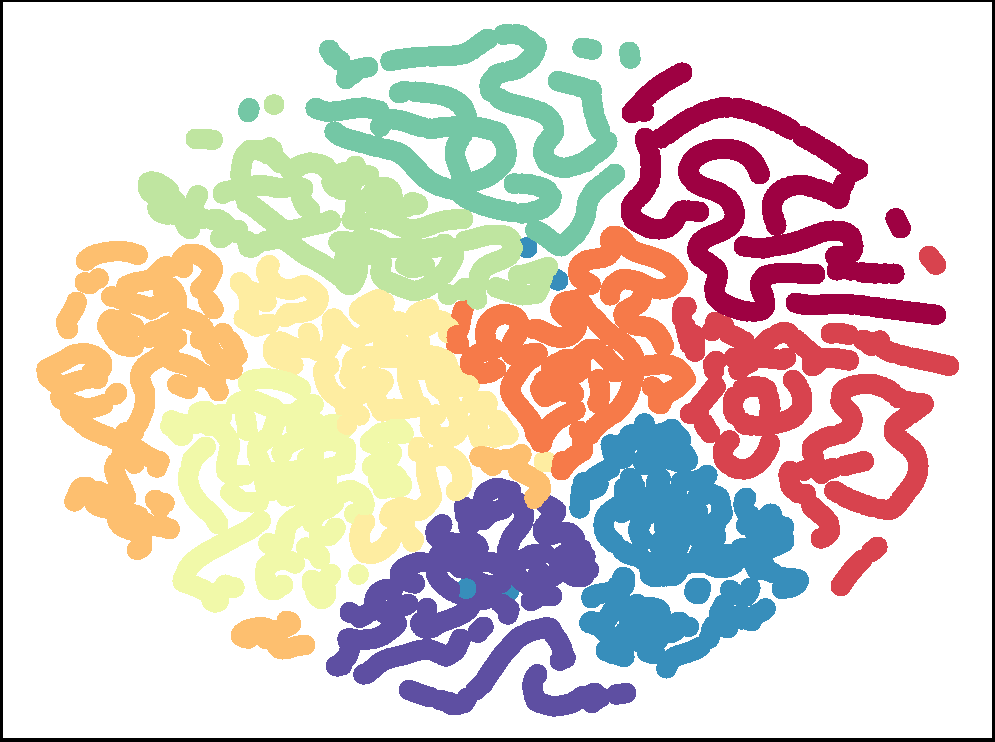}}}
\caption{t-SNE visualization of representations learned by SPC-MGR (a), SimMC (b), and TranSG (c) for first ten classes in IAS. Different colors indicates representations of different classes.}
\label{TSN_comp}
\end{figure}

\textbf{Discussions.} As presented in Fig. \ref{parameters}, we show effects of different hyper-parameters on our approach. An appropriate fusion of different components can encourage better model performance, while setting $\alpha$, $\beta$, and $\lambda$ to 0.5 achieves slightly better results. Adding too many FR heads or SGT layers could slightly reduce the performance, as it might expand the model scale and learn more redundant information. Our model is not sensitive to the variation of some parameters such as temperature $\tau_{1}$, while setting a moderate value for mask numbers benefits model performance. 
The t-SNE visualization \cite{van2008visualizing} in Fig. \ref{TSN_comp} shows that our learned skeleton representations possess more discriminative inter-class separation than other methods \cite{rao2022skeleton,rao2022simmc}, which indicates that TranSG may capture richer class-related semantics. 
More empirical and theoretical analyses are in the appendices.

\section{Conclusion}
In this paper, we propose TranSG to learn effective representations from skeleton graphs for person re-ID. We devise a skeleton graph transformer (SGT) to perform full-relation learning of body-joint nodes to aggregate key body and motion features into graph representations. A graph prototype contrastive learning (GPC) approach is proposed to learn discriminative graph representations by contrasting their inherent similarity with the most representative graph features. Furthermore, we design a graph structure-trajectory prompted reconstruction (STPR) mechanism to encourage learning richer graph semantics and key patterns for person re-ID. TranSG outperforms existing state-of-the-art models, and can be scalable to be applied to different scenarios.


\section*{Acknowledgements}
This research is supported by the National Research Foundation, Singapore under its AI Singapore Programme (AISG Award No: AISG2-PhD/2022-01-034[T]).

{\small
\bibliographystyle{ieeetr}
\bibliography{main}
}
\end{document}